\pdfoutput=1

\documentclass[11pt]{article}

\usepackage[final]{acl}

\usepackage{times}
\usepackage{latexsym}

\usepackage[T1]{fontenc}

\usepackage[utf8]{inputenc}

\usepackage{microtype}

\usepackage{inconsolata}

\usepackage{graphicx}

\usepackage{url}
\usepackage{amsmath}
\usepackage{amssymb}
\usepackage{multicol}
\usepackage{multirow}

\title{\textit{Yeah}, \textit{Un}, \textit{Oh}: Continuous and Real-time Backchannel Prediction\\with Fine-tuning of Voice Activity Projection}

\author{
 \textbf{Koji Inoue},
 \textbf{Divesh Lala},
 \textbf{Gabriel Skantze\textsuperscript{*}},
 \textbf{Tatsuya Kawahara}
\\
\\
 Graduate School of Informatics, Kyoto University, Japan, \\
 \textsuperscript{*}KTH Royal Institute of Technology, Sweden
\\
 \small{
   \textbf{Correspondence:} \href{mailto:inoue@sap.ist.i.kyoto-u.ac.jp}{inoue@sap.ist.i.kyoto-u.ac.jp}
 }
}

\newcommand{\ph}[1]{\phantom{#1}}

\begin{document}
\maketitle

\begin{abstract}
In human conversations, short backchannel utterances such as ``yeah'' and ``oh'' play a crucial role in facilitating smooth and engaging dialogue.
These backchannels signal attentiveness and understanding without interrupting the speaker, making their accurate prediction essential for creating more natural conversational agents.
This paper proposes a novel method for real-time, continuous backchannel prediction using a fine-tuned Voice Activity Projection (VAP) model.
While existing approaches have relied on turn-based or artificially balanced datasets, our approach predicts both the timing and type of backchannels in a continuous and frame-wise manner on unbalanced, real-world datasets.
We first pre-train the VAP model on a general dialogue corpus to capture conversational dynamics and then fine-tune it on a specialized dataset focused on backchannel behavior.
Experimental results demonstrate that our model outperforms baseline methods in both timing and type prediction tasks, achieving robust performance in real-time environments.
This research offers a promising step toward more responsive and human-like dialogue systems, with implications for interactive spoken dialogue applications such as virtual assistants and robots.
\end{abstract}

\renewcommand{\thefootnote}{\fnsymbol{footnote}}
\footnote[0]{This paper has been accepted for presentation at the main conference of 2025 Annual Conference of the Nations of the Americas Chapter of the Association for Computational Linguistics (NAACL 2025) and represents the author's version of the work.}
\renewcommand{\thefootnote}{\arabic{footnote}}

\section{Introduction}

In natural human conversations, short backchannels, such as ``{\it yeah}'' and ``{\it right},'' play a crucial role in facilitating smooth and engaging interactions~\cite{clark1996using,clancy1996conversational,maddrell2012influence}.
They function as feedback mechanisms, signaling attentiveness, understanding, and agreement without interrupting the speaker.
Accurate prediction and generation of backchannels in spoken dialogue systems are essential for creating more natural and human-like interactions~\cite{schroder2011building,devault2014simsensei,inoue2020sigdial}.
Although some definitions of backchannels include longer and more linguistic tokens such as ``{\it I see},'' this work focuses on short tokens that are frequently and dynamically used by listeners.

\begin{figure}[t]
    \includegraphics[width=\linewidth]{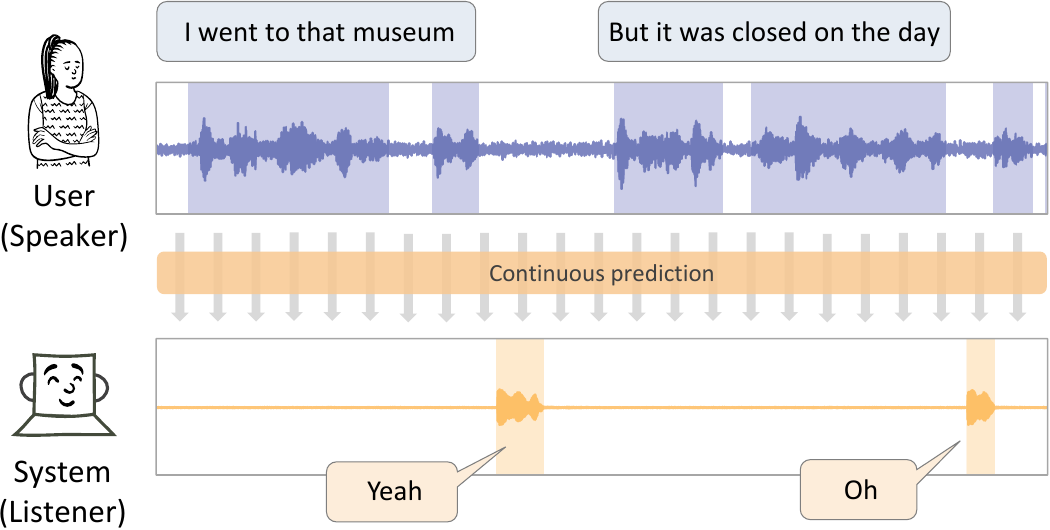}
    \caption{Conceptual diagram of continuous backchannel prediction}
    \label{fig:img:bc}
\end{figure}

Backchannel modeling remains a significant challenge due to their subtle and context-dependent characteristics.
Given the dynamic nature of backchannels, it is essential to predict them on a frame-by-frame basis and in real-time for live spoken dialogue systems, as depicted in Figure~\ref{fig:img:bc}.
However, most previous studies have focused on utterance-based systems, or, in the case of frame-based systems, have artificially balanced the test data by reducing non-backchannel samples.
This data manipulation introduces a discrepancy between the training models and real-world systems. 
Consequently, for practical applications, it is necessary to develop models capable of real-time, continuous frame-wise prediction and evaluate them using unbalanced, real-world datasets.

Transformer-based architectures have emerged as powerful tools for a broad range of sequential prediction tasks, such as language modeling and speech recognition. Among these, the Voice Activity Projection (VAP) model, a Transformer-based architecture, has shown its efficacy in predicting future voice activity within dialogues~\cite{erik2022vap,erik2022sigdial}.
Since voice activity is closely intertwined with turn-taking dynamics and backchannel behaviors, the VAP model seems to have the potential to enable more accurate backchannel prediction.
Furthermore, as a previous study implemented the real-time VAP model~\cite{inoue2024iwsds}, a VAP-based backchannel prediction model could serve as a promising candidate for real-time backchannel generation systems.
In its initial application, the VAP model was employed for backchannel prediction in a zero-shot manner.
However, this approach exhibited a sensitivity to threshold selection and depended on a balanced dataset.
These findings suggest that while the VAP model shows promise for backchannel prediction, further research is necessary to refine its training methodology for improved performance.

In this study, we propose a novel approach to backchannel prediction by utilizing the VAP model as a foundational framework.
We first train the VAP model on a large corpus of general dialogue data to capture the fundamental patterns of conversational dynamics.
Subsequently, we fine-tune the model on a specialized dataset focused on backchannel prediction.
This two-stage training process is analogous to the pre-training and fine-tuning paradigm employed by models like BERT, aiming to demonstrate the VAP model's versatility as a general-purpose base model.
Moreover, to the best of our knowledge, our model is the first to predict both the timing and type of backchannels in a continuous and real-time manner.

The contributions of this paper are twofold.
\begin{itemize}
  \setlength{\itemsep}{0pt}
  \setlength{\parskip}{0pt}
  \setlength{\parsep}{0pt}
    \item \textbf{Real-Time Continuous Backchannel Prediction}: A method for real-time, continuous backchannel prediction based on the VAP model is developed and evaluated on real-world, unbalanced test data.
    \item \textbf{Two-Stage Training for Generalization}: A two-stage training process is introduced for the VAP model, demonstrating its potential as a fundamental model for predicting conversational dynamics.
\end{itemize}
Note that the source codes and trained models are publicly available~\footnote{We will put the link here after the reviewing process.}.

\section{Related Work}

Effective backchannel generation necessitates accurate prediction of three key elements: temporal placement (timing), linguistic form (type), and prosodic patterns.
The majority of prior studies have focused exclusively on predicting the timing and types of backchannels.
The definition and functions of backchannel types have been explored in conversational analysis and linguistic studies~\cite{drummond1993back,wong2007study,tang2009contrastive,den2011annotation}.
Despite the critical role of prosody in entrainment, existing research on this remains scarce~\cite{kawahara2015iwsds,ochi2024interspeech}.
This review primarily summarizes the current state of research on predicting the timing and form of backchannels.

Before recent advances in machine learning technologies, backchannel prediction models were primarily based on hand-crafted features with heuristic rules or simpler models~\cite{koiso1998analysis,ward2000prosodic,fujie2005back,morency2008predicting,morency2010probabilistic,ozkan2011modeling,blache2020integrated}.
With the advent of dataset creation paradigms and machine learning advancements, deep learning models began to be employed for backchannel prediction. 
Initial models were built using long-short-term memory (LSTM) networks~\cite{ruede2017enhancing,ruede2017yeah,adiba2021towards,jain2021exploring}, while more recent approaches leverage Transformer-based models~\cite{jang2021bpm_mt,liermann2023emnlp}.

Most previous studies focused on timing prediction, with some also addressing type prediction. 
The most conventional approach involves framing the prediction task as a three-class classification problem: non-backchannel, continuer, and assessment~\cite{choi2024etri}, as described in Section~\ref{def:multi}. 
The present work adopts this three-class classification scheme for both timing and type prediction. 
Other research has explored a four-type classification system, encompassing non-backchannel, continuer, understanding, and empathy backchannels~\cite{jang2021bpm_mt}. 
Another study introduced a five-type classification for single continuer, double continuers (e.g., "yeah yeah"), triple continuers, assessment, and non-backchannel~\cite{kawahara2016interspeech}. 
Furthermore, a different approach proposed a two-step classification method where the first model predicts the timing, followed by a second model that determines the type~\cite{adiba2021towards}. 

In terms of prediction unit, utterance-based or continuous, utterance-based models tend to incorporate linguistic features such as word embeddings~\cite{jang2021bpm_mt,park2024improving}. 
Conversely, previous continuous models were generally restricted to using prosodic features~\cite{ruede2017enhancing,ruede2017yeah}. 
Recent models have begun utilizing audio encoders that can theoretically capture both linguistic and prosodic information in an end-to-end manner~\cite{park2024improving,choi2024etri}. 
The VAP model used in this study similarly employs a pre-trained contrastive predictive coding (CPC) model as its audio encoder. 

To address the issue of imbalanced data, recent studies have integrated multi-task learning with the primary task of backchannel prediction. 
For instance, subtasks such as turn-taking prediction~\cite{hara2018multi,ishii2021multimodal}, sentiment score analysis~\cite{jang2021bpm_mt}, dialogue act recognition~\cite{liermann2023emnlp}, and streaming automatic speech recognition (ASR)~\cite{choi2024etri} have been considered. 
Notably, a recent model~\cite{choi2024etri} was evaluated on a real-world imbalanced dataset, demonstrating reasonable performance with F1-scores of 26\% and 22\% for continuer and assessment backchannels, respectively, in a frame-wise manner. 
This work also proposes the incorporation of multi-task learning for both backchannel prediction and VAP tasks. 

\section{Dataset}

We employ two types of datasets: one specialized for backchannel prediction and the other for pre-training the proposed model.
Note that all the dialogue datasets were in Japanese.

\begin{figure}[t]
    \includegraphics[width=\linewidth]{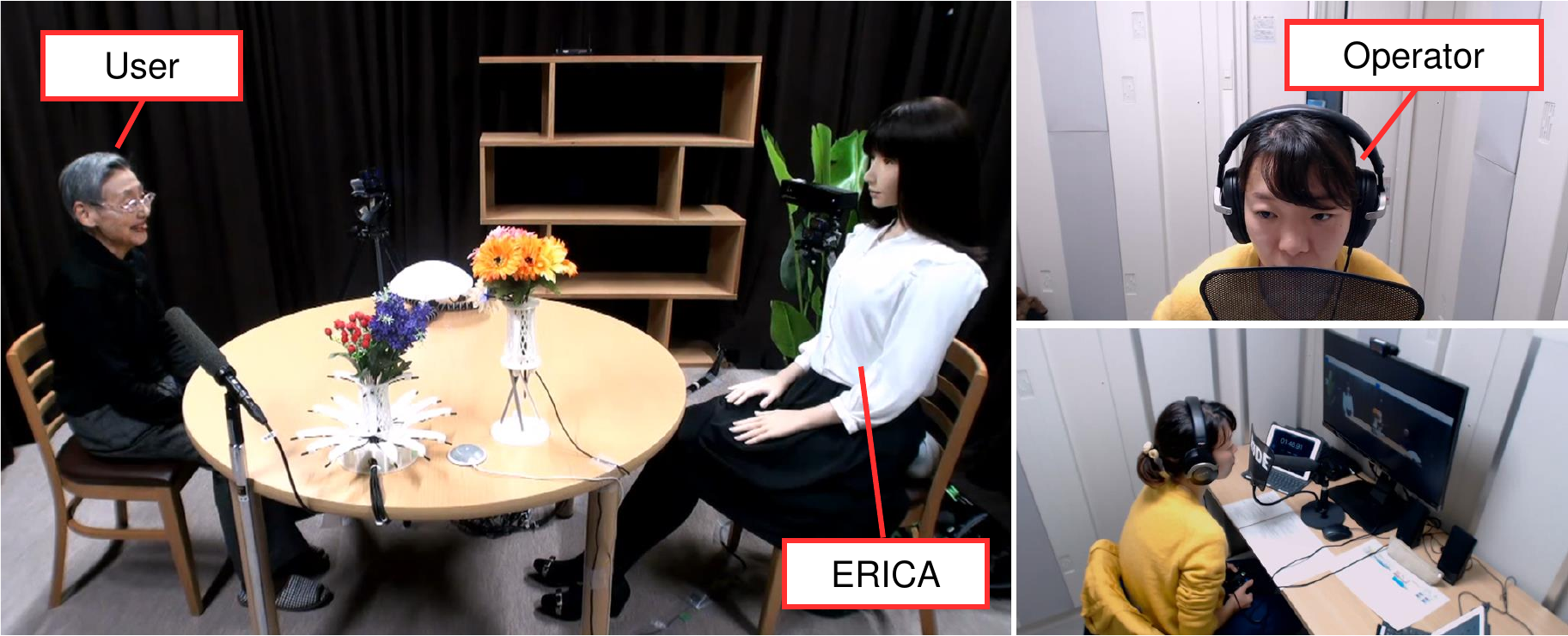}
    \caption{Setup for dialogue recording}
    \label{fig:erica}
\end{figure}

\subsection{Attentive Listening Dataset} \label{data:bc}

We have collected spoken dialogue data using a Wizard-of-Oz (WOZ) setup.
In this experiment, the android ERICA~\cite{inoue2016erica} was employed, with a human operator remotely controlling it, which was transmitted and played through ERICA’s speaker system (Figure~\ref{fig:erica}).

The dialogue task focused on attentive listening, where human participants (speaker) shared personal experiences, and ERICA actively engaged as a listener~\cite{inoue2020sigdial,lala2017sigdial}.
This task was advantageous because it allowed for the collection of numerous backchannel responses by ERICA.
The participants comprised two demographic groups: students and older adults.
Each group was provided with a prompt to guide their conversation; for instance, students discussed ``challenges during the COVID-19 pandemic,'' while the elderly participants reflected on ``memorable travel experiences and recent favorite meals.''

ERICA's operators were three actresses, who had experiences of our past attentive listening dialogue experiments.
While backchannel behaviors can be subjective and vary among individuals, the few operators were selected for their experience, ensuring the collection of high-quality backchannel data.
Furthermore, the operators participated in a sufficient number of dialogue sessions in this experiment, ensuring both the quantity and quality of the training data, despite the varying speakers in each session.

In total, we recorded 109 dialogue sessions, each lasting approximately 7 to 8 minutes.
The data were randomly divided into 87, 11, and 11 sessions for training, validation, and testing purposes, respectively.
We subsequently transcribed the dialogues and annotated ERICA's backchannel responses.

\subsection{Pre-training Data for VAP} \label{data:large}

In this study, we introduce a two-step training approach where the original VAP model is initially trained, followed by fine-tuning for backchannel prediction.
The first step requires a larger dataset to train the VAP model effectively to continuously predict future voice activities.
To support this, in addition to the attentive listening dialogue dataset mentioned earlier, we incorporated additional training data at this stage.
Using the same configuration as ERICA, we recorded data across various scenarios, such as job interviews~\cite{inoue2020icmi} and first-meeting dialogues~\cite{inoue2022laugh}.
These diverse tasks provide different dialogue styles, enhancing the VAP model's robustness and enabling it to adapt to various behaviors, including backchannels.
In total, the pre-training data amounted to about 35 hours, which includes the training set from the aforementioned backchannel prediction dataset.

\section{Task Definition}

In this work, we address two distinct backchannel prediction tasks as outlined below.
 
\subsection{Timing Prediction} \label{def:binary}

\begin{figure}[t]
    \includegraphics[width=\linewidth]{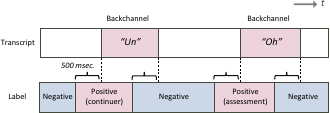}
    \caption{Definition of positive (backchannel) and negative (non-backchannel) frames}
    \label{fig:def:bc}
\end{figure}

The primary objective of this task is to predict the occurrence of a backchannel, framing it as a binary classification problem.
We manually annotated the backchannels in the aforementioned dialogue dataset, identifying two distinct types of short tokens as backchannels: continuer and assessment.
The continuer tokens include expressions such as ``{\it un}'' and ``{\it hai}'' in Japanese, which correspond to ``{\it yeah}'' and ``{\it right}'' in English.
On the other hand, the assessment tokens include utterances such as ``{\it he-}'' and ``{\it oh}'' in Japanese, equivalent to ``{\it wow}'' and ``{\it oh}'' in English.
It is important to note that in the current task, we do not differentiate between these two token types, whereas such a distinction is made in the second task.
To facilitate the implementation of the model in real-time spoken dialogue systems, we marked the positive sample frames as occurring 500 milliseconds before the actual backchannel utterances, as illustrated in Figure~\ref{fig:def:bc}.

The total number of annotated backchannel utterances amounted to 13,601, with a cumulative duration of 5,912.6 seconds.
These were split into 11,371 utterances for training, 1,139 for validation, and 1,091 for testing.
For the negative samples, the cumulative time of non-backchannel segments was 56,467.3 seconds, resulting in a ratio of approximately 10\% for positive samples.

\subsection{Timing and Type Prediction} \label{def:multi}

In the second task, the prediction process becomes more refined by incorporating different types of backchannels. 
Although numerous definitions of backchannel types or categories exist in prior research, we adopt the two basic types: continuers and assessments, as defined earlier. 
This distinction is crucial for conveying different listener intentions, and most previous studies have primarily addressed continuers, as assessment backchannels may require comprehension of both the prosodic and linguistic aspects of the user's utterances. 

After reclassifying the previously annotated backchannels into these two categories, we found that there were 10,081 instances of continuers and 3,506 instances of assessment backchannels.
This ratio means that the prediction of assessments seems to be more difficult than those for continuers.
It is important to note that 14 tokens could not be classified into either category, and thus they were excluded from this task. 
Consequently, the classification problem in this task becomes a three-class classification: continuers, assessments, and non-backchannels. 
The definition of timing remains consistent with the previous task, as illustrated in Figure~\ref{fig:def:bc}. 

\begin{figure}[t]
    \includegraphics[width=\linewidth]{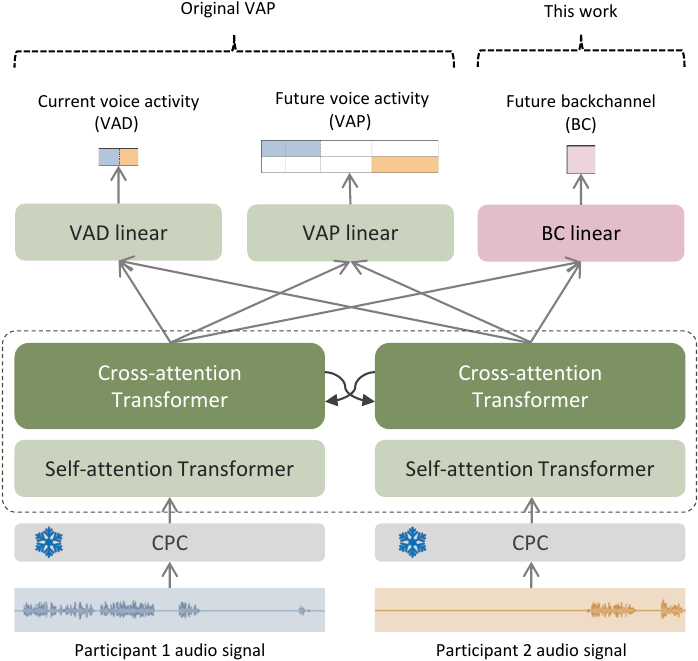}
    \caption{Architecture of the VAP model}
    \label{fig:vap}
\end{figure}

\section{Proposed Method}

In this section, we begin by explaining the voice activity projection (VAP) model, which serves as the foundational model and is trained to predict future voice activities by using the largest spoken dialogue dataset explained in Section~\ref{data:large}.
Following that, we discuss how to adapt a pre-trained VAP model for use in the current backchannel prediction task, using the data introduced in Section~\ref{data:bc}.

\subsection{Voice Activity Projection}

The VAP model employed in this study is constructed upon a Transformer-based architecture designed to emulate human-like predictive capabilities.
As illustrated in Figure~\ref{fig:vap}, the architecture of the VAP model processes stereo audio signals originating from two participants\footnote{A detailed explanation of the model's architecture is provided in a previous work~\cite{inoue2024coling}}, consistent with the operation of full-duplex spoken dialogue systems.
Note that this model integrates the listener's audio as one of the input channels.
This allows self-generated backchannels to be fed back into the model, distinguishing it from other existing models.
This functionality plays a crucial role in preventing multiple consecutive backchannels, which may appear unnatural.

Each audio channel is initially processed independently through a Contrastive Predictive Coding (CPC) audio encoder and a channel-wise Transformer.
The CPC was pre-trained with the Librispeech dataset~\cite{riviere2020unsupervised} and is frozen during the VAP training.
The resulting outputs are then input into a cross-attention Transformer, where one channel serves as the query, while the other functions as the key and value.
The output, a concatenation of both channels, produces a 512-dimensional vector.
Note that we used the same parameters as defined in the original work~\cite{inoue2024coling} where the numbers of layers for the channel-wise and the cross-attention Transformers were 1 and 3, and the number of attention heads was 4, respectively.

\begin{figure}[t]
    \includegraphics[width=\linewidth]{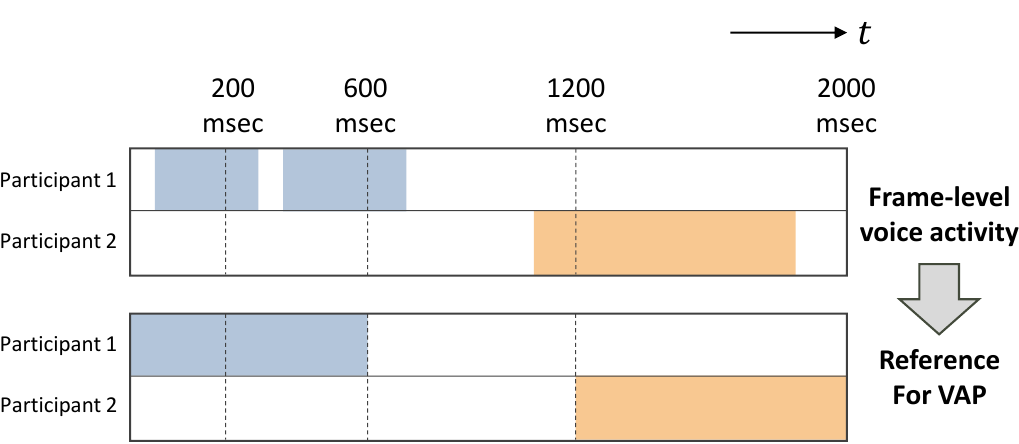}
    \caption{Definition of the VAP state}
    \label{fig:vap-ref}
\end{figure}

The concatenated vector is subsequently processed by linear layers for two distinct tasks: voice activity projection (VAP) and voice activity detection (VAD).
The primary task, voice activity projection, yields a 256-dimensional state vector that predicts the voice activity of the two participants over the next two seconds~\cite{erik2022vap}, as illustrated in Figure~\ref{fig:vap-ref}.
This two-second period is divided into four time intervals: 0-200 ms, 200-600 ms, 600-1200 ms, and 1200-2000 ms.
Consequently, there are eight binary bins in total, four for each participant, resulting in 256 ($=2^8$) possible combinations of speaking/non-speaking states for each participant within these bins.
The voice activity detection task, conversely, focuses on the current frame, producing two binary vectors representing the present voice activity of each participant.
Both tasks are trained using the cross-entropy loss function, denoted as $L_{vap}$ and $L_{vad}$ in the following subsection.

\subsection{Fine-Tuning for Backchannel Prediction}

Following the pre-training of the VAP model, an additional training phase is conducted using data specific to backchannel prediction.
To facilitate this, a new linear layer is introduced on top of the VAP model, complementing the existing layers for VAP and VAD, as depicted in Figure~\ref{fig:vap}.
The loss function for this fine-tuning process, denoted as $L$, is formulated as follows:
\begin{equation}
L = \alpha ~ L_{vap} + \beta ~ L_{vad} + \gamma L_{bc}~ , \label{loss}
\end{equation}
where $\alpha$, $\beta$, and $\gamma$ are the hyperparameters used to adjust the balance between the three tasks, with $\gamma$ typically assigned a higher value due to the primary focus on backchannel prediction.
The first two terms are consistent with those used in the original VAP model~\cite{erik2022vap}, while the final term, $L_{bc}$, is newly introduced in this work.
This term represents the cross-entropy loss associated with backchannel prediction and is defined as:
\begin{equation}
L_{bc} = - \log \sigma(\mathbf{o}_{bc}(r_{bc})) ~ , \\
\end{equation}
where $\mathbf{o}_{bc}$ represents the output from the linear head associated with backchannel prediction, and $r_{bc}$ denotes the reference label.
It is important to note that the dimensionality of these vectors is dependent on the specific task.
For instance, in binary classification tasks, such as predicting the presence or absence of a backchannel (Section~\ref{def:binary} and Section~\ref{exp:binary}), the dimensionality would be 2.
Conversely, in multi-class classification tasks, which involve predicting both the timing and type of backchannel (Section~\ref{def:multi} and Section~\ref{exp:multi}), the dimensionality would exceed 3.

\begin{figure*}[t]
    \includegraphics[width=\linewidth]{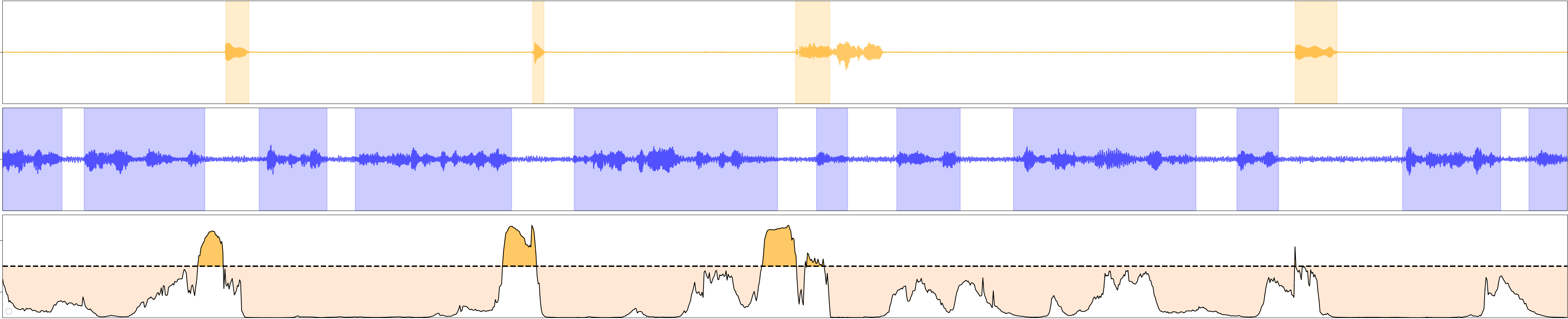}
    \caption{Example of backchannel timing prediction. The top section represents the listener's activity (orange: backchannel), the middle shows the speaker's speech, and the bottom section illustrates the model's predicted probabilities for backchannel occurrence.}
    \label{fig:result:vis1}
\end{figure*}

\section{Experiment}

To evaluate the effectiveness and applicability of the proposed method, we conducted the four experiments described below.

\subsection{Timing Prediction} \label{exp:binary}

The first experiment focuses on the initial tasks outlined in Section~\ref{def:binary}, which involve a binary classification of backchannels or non-backchannels. 
We prepared several comparative methods, including random classification (always outputs positive), as detailed below:
\begin{itemize}
    \item[(i)] \textbf{Baseline} consists solely of the audio encoder CPC and a linear head. 
    While other methods in this study freeze the CPC during training, this baseline approach fine-tunes the CPC model itself. 
    As the two input audio channels are separately fed into the CPC, their distinct output vectors are concatenated and then passed to the linear head.
    \item[(ii)] \textbf{Zero-shot} is based on the direct use of the VAP output proposed in the original VAP work~\cite{erik2022vap}.
    It is a normalized value obtained by adding the probability of immediate voice activity of the listener (system) (0.0 to 1.2 seconds) and the probability of slightly later voice activity of the speaker (user) (1.2 to 2.0 seconds).
    A higher value indicates that the immediate system utterance would be short, corresponding to a backchannel.
    \item[(iii)] \textbf{ST w/o PT} refers to the single-task (ST) model, where the loss function only includes $L_{bc}$ from Equation~(\ref{loss}).
    Moreover, this model does not involve any pre-training (PT) of the VAP model.
    \item[(iv)] \textbf{ST w/ PT} introduces pre-training of the VAP model in addition to the single-task learning.
    \item[(v)] \textbf{MT-ASR} performs another type of multi-task learning, where backchannel prediction is trained together with automatic speech recognition, inspired by a recent backchannel prediction model~\cite{choi2024etri}.
    This subtask was trained to recognize phonemes (19 phones) using the CTC loss function.
    \item[(vi)] \textbf{MT w/ PT} represents the proposed method, which incorporates both multi-task (MT) learning, as described in Equation~\ref{loss}, and the pre-training of the VAP model.
\end{itemize}
The evaluation metrics used are F1-score, precision, and recall, calculated in a frame-wise manner, and the F1-score is the most priority indicator.
For the hyperparameters in Equation~(\ref{loss}), to emphasize the importance of backchannel prediction, we empirically set them as $\alpha=1$, $\beta=1$, and $\gamma=5$.
Additionally, to mitigate the impact of the imbalanced dataset, we adjusted the loss weight, assigning a weight five times larger to positive (backchannel) samples compared to negative (non-backchannel) samples.

\begin{figure*}[t]
    \includegraphics[width=\linewidth]{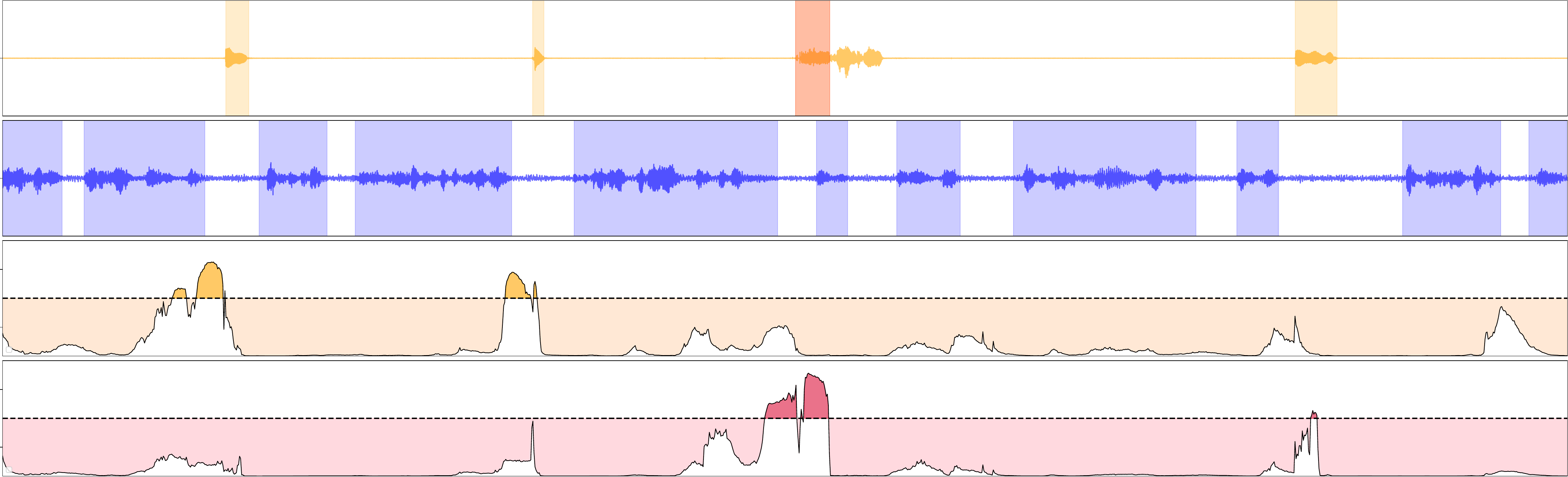}
    \caption{Example of model predictions for backchannel timing and type. The top two sections show the listener (orange: continuer, pink: assessment) and speaker activities. The third section displays the model's prediction probabilities for continuer (orange), while the bottom section shows the probabilities for assessment (pink).}
    \label{fig:result:vis2}
\end{figure*}

\begin{table}[t]
    \centering
    \caption{Result on timing prediction (ST: Single-task, MT: Multi-task, PT: Pre-training)}
    \begin{tabular}{lccc}
        \hline
        \multicolumn{1}{c}{Method} & F1-score & Precision & Recall \\
        \hline
        Random & 13.76 & \ph{0}7.39 & 100.00 \\
        Zero-shot & 15.11 & \ph{0}8.22 & \ph{0}93.11 \\
        Baseline & 36.37 & 26.43 & \ph{0}58.32 \\
        ST w/o PT & 36.34 & 25.04 & \ph{0}66.24 \\
        ST w/ PT & 41.65 & 31.31 & \ph{0}62.18 \\
        MT-ASR & 39.34 & 28.25 & 63.39 \\
        MT w/ PT & 42.85 & 32.52 & \ph{0}62.80 \\
        \hline
    \end{tabular}
    \label{tab:result:timing}
\end{table}

Table~\ref{tab:result:timing} presents a summary of the results for this task. 
Firstly, all trained methods demonstrated significantly higher scores when compared to the random and zero-shot approaches.
In particular, the proposed method (MT w/ PT) achieved the highest scores in both the F1 score and precision metrics.
These findings suggest that both pre-training and multi-task learning play a crucial role in improving backchannel prediction performance, indicating that this task requires a more robust approach than conventional target-specific training or fine-tuning. 
Moreover, the VAP model, along with its original loss function, exhibits better generalizability and applicability to other non-linguistic behavior predictions, such as the current backchannel prediction task.

Figure~\ref{fig:result:vis1} illustrates a sample output generated by the model.
As shown in the graph, even while the Blue speaker is still speaking, the model is capable of predicting multiple backchannel points just prior to their occurrence.

\begin{table}[t]
    \centering
    \caption{Result on time and type prediction on \textbf{continuer} (ST: Single-task, MT: Multi-task, PT: Pre-training)}
    \begin{tabular}{lccc}
        \hline
        \multicolumn{1}{c}{Method} & F1-score & Precision & Recall \\
        \hline
        Random & 10.19 & \ph{0}5.37 & 100.00 \\
        Baseline & 34.13 & 26.59 & \ph{0}47.63 \\
        ST w/o PT & 36.10 & 28.65 & \ph{0}48.77 \\
        ST w/ PT & 36.47 & 29.08 & \ph{0}48.90 \\
        MT-ASR & 33.45 & 25.33 & \ph{0}49.26 \\
        MT w/ PT & 38.11 & 29.89 & \ph{0}52.58 \\
        \hline
    \end{tabular}
    \label{tab:result:type:con}
\end{table}

\subsection{Timing and Type Prediction} \label{exp:multi}

The second experiment involves the prediction of backchannel types, as outlined in Section~\ref{def:multi}.  
We employed the same comparative methods as in the previous experiment, but adjusted the output dimension of the linear head from 2 to 3 to accommodate the classification of continuers, assessments, and non-backchannels.  
The evaluation metric remained unchanged; however, we conducted separate evaluations for continuers and assessments.
Note that the zero-shot approach (defined in Section~\ref{exp:binary}) was not applicable to this task and was therefore excluded.

Table~\ref{tab:result:type:con} and Table~\ref{tab:result:type:ass} present the outcomes of this task.
As with the previous results, both tables demonstrate that the combination of multi-task learning and pre-training significantly enhanced performance, with the proposed method (MT w/ PT) achieving the highest F1-score.
When comparing these two types, as anticipated, the prediction of assessment backchannels yielded lower scores.
While random prediction offers no meaningful insight, the proposed method exceeded an F1-score of 30.

Figure~\ref{fig:result:vis2} illustrates a sample output generated by the model.
In this example, the listener uttered two continuer backchannels (orange) followed by an assessment (pink).
The model can correctly predict the first two continues and then also properly predict the assessment.
From this result, the model can be trained properly to predict both two types of backchannels.

\begin{table}[t]
    \centering
    \caption{Result on time and type prediction on \textbf{assessment} (ST: Single-task, MT: Multi-task, PT: Pre-training)}
    \begin{tabular}{lccc}
        \hline
        \multicolumn{1}{c}{Method} & F1-score & Precision & Recall \\
        \hline
        Random & \ph{0}3.57 & \ph{0}1.82 & 100.00 \\
        Baseline & 19.74 & 32.71 & \ph{0}14.13 \\
        ST w/o PT & 23.72 & 26.11 & \ph{0}21.73 \\
        ST w/ PT & 30.09 & 30.36 & \ph{0}29.82 \\
        MT-ASR & 20.27 & 25.33 & \ph{0}16.90 \\
        MT w/ PT & 31.76 & 29.95 & \ph{0}33.81 \\
        \hline
    \end{tabular}
    \label{tab:result:type:ass}
\end{table}


\subsection{Prosody Sensitivity}

\begin{figure}[t]
  \begin{minipage}{0.49\linewidth}
    \centering
    Original (before)
    \includegraphics[width=\linewidth]{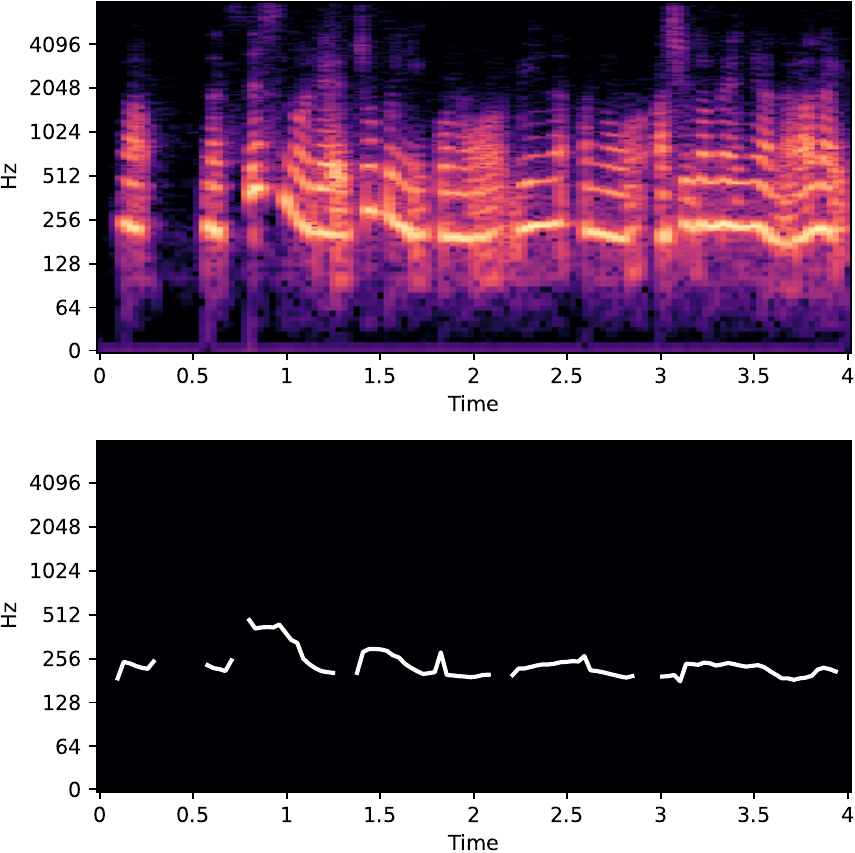}
  \end{minipage}
  \begin{minipage}{0.49\linewidth}
    \centering
    Pitch flattened (after)
    \includegraphics[width=\linewidth]{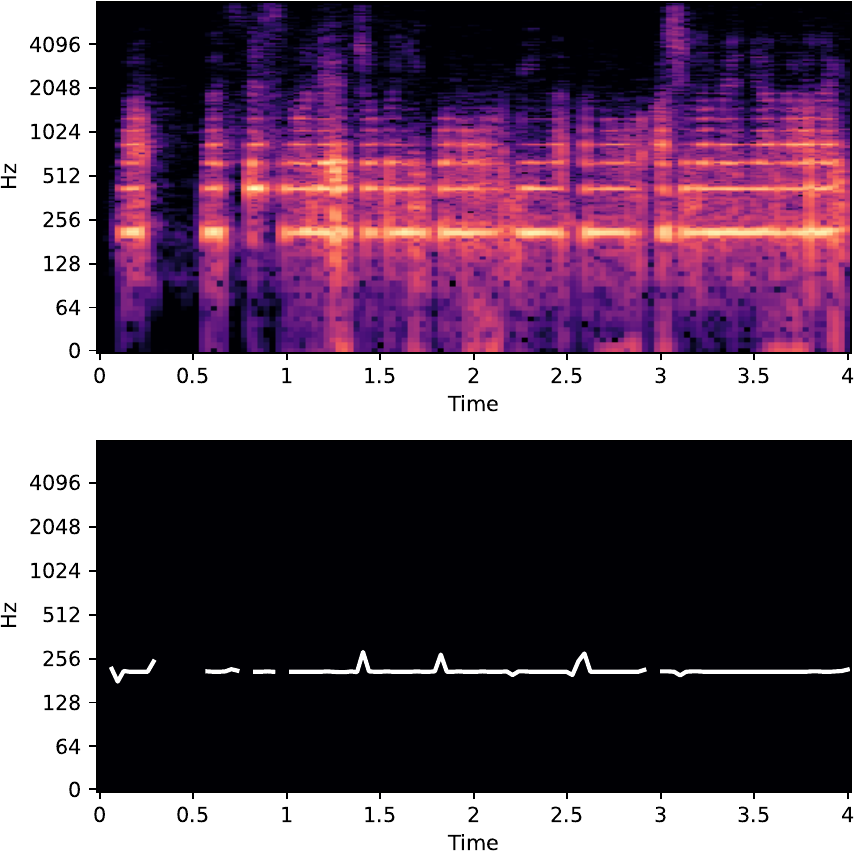}
  \end{minipage}
  \caption{Input example of pitch flattening test (Top: Spectrogram, Bottom: Automatically estimated F0)}
  \label{fig:exp:pitch_flatten}
\end{figure}

\begin{figure}[t]
  \begin{minipage}{0.49\linewidth}
    \centering
    Original (before)
    \includegraphics[width=\linewidth]{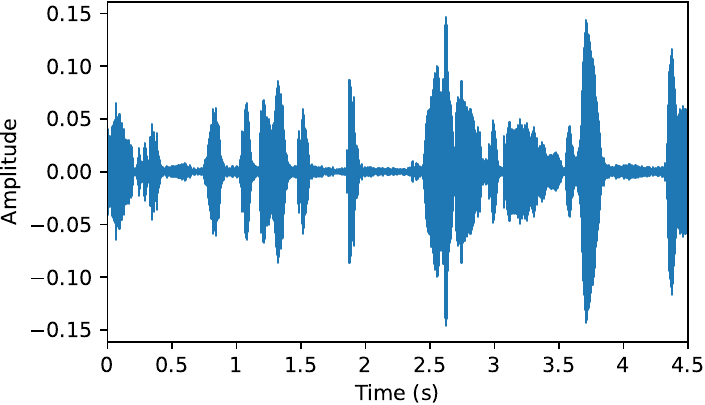}
  \end{minipage}
  \begin{minipage}{0.49\linewidth}
    \centering
    Intensity flattened (after)
    \includegraphics[width=\linewidth]{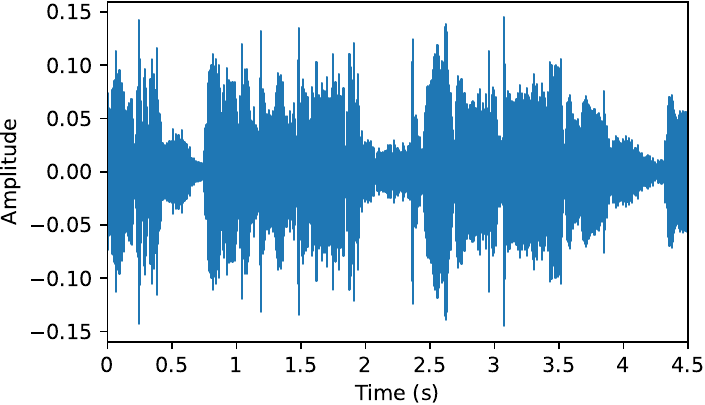}
  \end{minipage}
  \caption{Input example of intensity flattening test}
  \label{fig:exp:intensity_flatten}
\end{figure}

We further examined the extent to which the model depends on prosodic information.
Previous work on the VAP model conducted a similar experiment by flattening the pitch (Figure~\ref{fig:exp:pitch_flatten}) and intensity (Figure~\ref{fig:exp:intensity_flatten}) of the test input~\cite{erik2022sigdial}.
In this study, we similarly utilized Praat\footnote{\url{https://www.fon.hum.uva.nl/praat/}} to flatten both pitch and intensity, respectively.
If such manipulations significantly degrade performance, it would suggest that the model both relies on and effectively captures the prosodic information.
We subsequently analyzed the performance changes before and after applying the flattening manipulations across the three classification models.
Due to space limitations, we report only the F1-score of the proposed model (\textbf{MT w/PT}).

\begin{table}[t]
    \centering
    \caption{Pitch and intensity flattening result}
    \begin{tabular}{lcc}
        \hline
        \multicolumn{1}{c}{\multirow{2}{*}{Manipulation}} & \multicolumn{2}{c}{F1-score} \\
        \cline{2-3}
         & Continuer & Assessment \\
        \hline
        None (original)    & 38.11\phantom{ (-1.00)}   & 31.76\phantom{ (-1.00)} \\
        Pitch flat       & 37.20 (-0.91)             & 31.09 (-0.67) \\
        Intensity flat   & 35.48 (-2.63)             & 28.73 (-3.03) \\
        \hline
    \end{tabular}
    \label{tab:result:pitch}
\end{table}

Table~\ref{tab:result:pitch} presents the results of this experiment.
Overall, neither manipulation significantly degraded performance, suggesting that the model may rely more heavily on other factors, such as linguistic information.
Both types of backchannels exhibited a similar trend, with intensity flattening causing greater degradation than pitch flattening.
This finding indicates that the current backchannel prediction model captures the intensity dynamics of preceding user utterances more effectively.
When comparing the two types of backchannels, the assessment revealed a higher dependence on intensity, though the difference was not substantial.

\subsection{Real-time Processing Performance}

To validate the applicability of live spoken dialogue systems, we also examined the relationship between the model's input context length and its prediction performance.
As the CPC audio encoder is composed of an autoregressive model, we provided the entire context audio input to the encoder.
Subsequently, we constrained the input length for the Transformer layers.
In addition, we adjusted the frame rate to 10 Hz, which is sufficient for real-time prediction systems, and retrained the models accordingly.
Therefore, note that the results in this section would be different from the ones so far.
In this experiment, we employed the second task, which involves predicting both the timing and type of backchannels.
For this evaluation, only a CPU was utilized, specifically an Intel Core i7-11700 @ 2.50 GHz.

The result for the continuer and assessment backchannels in the different input context lengths is presented in Table~\ref{tab:result:real}.
Also, in this experiment, we only reported the F1-score of the proposed model (\textbf{MT w/PT}).
Overall, due to the compact design of the VAP model, the real-time factor (RTF) was consistently below 1.0 in all cases, indicating that real-time processing is achieved.
Regarding the effect of input context length on the Transformer layers, approximately 5 seconds of input context yielded optimal results for both types of backchannels.
When comparing the two types, while the continuer backchannels could be predicted even with a 1-second input context, the performance for assessment backchannels decreased significantly with shorter contexts, such as 1 or 3 seconds.
This disparity suggests that the prediction of assessment backchannels requires a longer input context.

\begin{figure*}[t]
    \includegraphics[width=\linewidth]{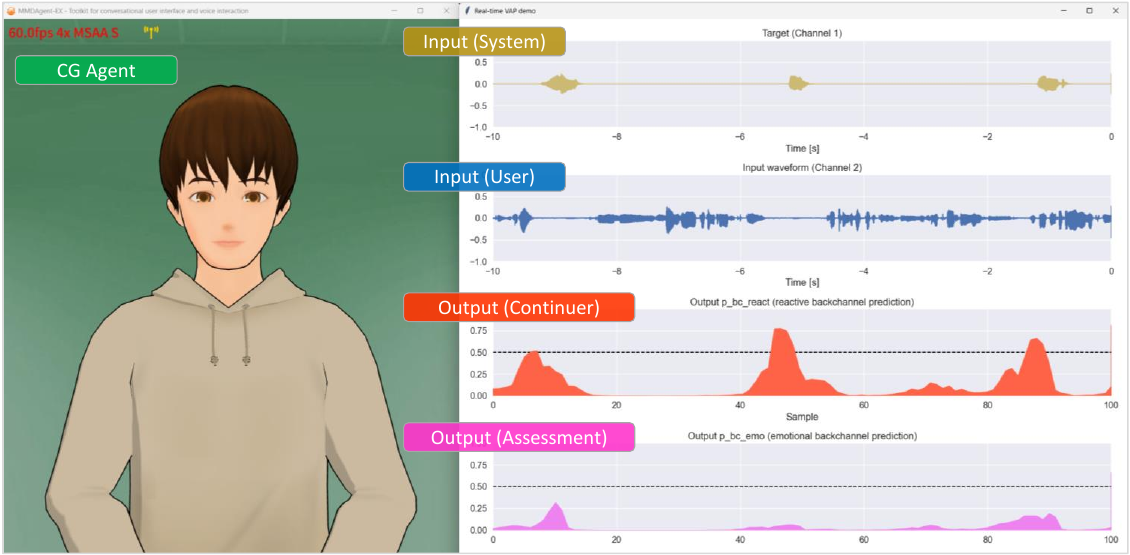}
    \caption{A conversational agent with VAP-based backchannel prediction and its GUI visualization}
    \label{fig:agent}
\end{figure*}

\begin{table}[t]
    \centering
    \caption{Real-time processing performance (RTF: Real-time factor)}
    \begin{tabular}{cccc}
        \hline
        \multirow{2}{*}{\shortstack{Context\\$[$sec.$]$}} & \multicolumn{2}{c}{F1-score} & \multirow{2}{*}{RTF} \\
        \cline{2-3}
         & Continuer & Assessment & \\
        \hline
        20 & 36.17 & 28.75 & 0.229 \\
        10 & 36.51 & 30.46 & 0.220 \\
        \ph{0}5 & 36.57 & 30.08 & 0.194 \\
        \ph{0}3 & 35.79 & 29.51 & 0.172 \\
        \ph{0}1 & 35.25 & 27.67 & 0.157 \\
        \hline
    \end{tabular}
    \label{tab:result:real}
\end{table}

\section{Integration with a CG Agent}

Based on the results from the previous experiment, we have developed a VAP-based real-time backchannel prediction system and implemented it with a conversational CG agent\footnote{CG-CA Takumi (c) 2023 by Nagoya Institute of Technology, Moonshot R\&D Goal 1 Avatar Symbiotic Society}.
Figure~\ref{fig:agent} illustrates the system in operation with the agent, as well as its real-time GUI visualization tool.
Note that since this backchannel generation system operates independently of the interface, it can be applied to various other interfaces, including physically embodied robots.
We plan to conduct a user dialogue experiment with this dialogue system to evaluate the naturalness and effectiveness of the backchannel generation system. 

\section{Conclusion}

This paper presents a method for real-time, continuous backchannel prediction using a fine-tuned Voice Activity Projection (VAP) model.
Our approach combines pre-training on a larger dialogue dataset with fine-tuning on a specialized backchannel dataset, leveraging the VAP architecture's generalizability. 
Experimental results showed that our two-stage and multi-task training process improves the model’s ability to predict both the timing and type of backchannels, demonstrating its adaptability to real-world, unbalanced data.
We also validated the model for real-time use, confirming its effectiveness in live systems without compromising accuracy, especially for continuer backchannels.
The results also highlight the need for longer input contexts for accurate assessment backchannel predictions.

This study represents a step forward in enhancing conversational agents’ interactivity by providing a more human-like and responsive backchanneling system.
Future research will concentrate on evaluating the effectiveness of the backchannel generation system through user dialogue experiments, as well as further refining backchannel prediction for more complex conversational contexts.




\section*{Acknowledgments}

This work was supported by JST PREST JPMJPR24I4, JST Moonshot R\&D JPMJPS2011, and JSPS KAKENHI JP23K16901.

\section*{Limitations}

This study was evaluated solely on a Japanese dialogue dataset, which limits the generalizability of the model to other languages.
Future work should assess its performance on common other datasets like Switchboard to ensure broader applicability.
Additionally, while our model shows promise for real-time backchannel prediction, it has not been evaluated in practical settings with conversational agents or robots.
Future experiments involving user interactions with such systems are necessary to evaluate the model’s effectiveness and user impressions in real-world scenarios.

\section*{Ethical Considerations}

In the process of collecting dialogue data, all participants were informed about the purpose of the research, and their explicit consent was obtained for the use of their data.
The data collection process was designed to ensure the protection of participants' privacy, and any personal information was anonymized or excluded from the dataset to prevent identification.
The study was conducted in accordance with ethical guidelines, and approval was obtained from the appropriate ethics committee prior to data collection.
The participants' privacy and confidentiality were strictly maintained throughout the research process.

\bibliography{acl_latex}




\end{document}